\PassOptionsToPackage{hyphens}{url}
\documentclass[letterpaper]{article}
\usepackage{spconf,amsmath,graphicx}
\usepackage{amssymb,mathtools,amsthm}
\usepackage{booktabs,multirow,array}
\usepackage[table]{xcolor}
\usepackage{subcaption}
\usepackage{listings}
\usepackage{cite}
\usepackage{xurl}
\usepackage{placeins}
\usepackage{balance}
\usepackage[hidelinks]{hyperref}

\usepackage{amsmath,amsfonts,bm}

\def\eqref#1{equation~\ref{#1}}

\def\1{\bm{1}}

\DeclareMathAlphabet{\mathsfit}{\encodingdefault}{\sfdefault}{m}{sl}
\SetMathAlphabet{\mathsfit}{bold}{\encodingdefault}{\sfdefault}{bx}{n}

\definecolor{downcolor}{HTML}{3A923A}
\definecolor{upcolor}{HTML}{B83945}

\renewcommand{\paragraph}[1]{\par\medskip\noindent\textbf{#1}\hspace{0.5em}\ignorespaces}

\hypersetup{
  pdftitle={HGDNA: Overlapping k-mer Tokenization for Efficient Autoregressive Genomic Language Models},
  pdfauthor={Xudong Wang; Siwei Cheng; Haozhe Hu;},
  pdfsubject={ICASSP 2027 format-only full-length working manuscript},
  pdfkeywords={genomic language models, overlapping k-mer tokenization, causal language modeling, linear attention, sequence classification}
}
\title{\fontsize{14}{16}\selectfont HGDNA: Overlapping k-mer Tokenization for Efficient Autoregressive Genomic Language Models}
\name{Xudong Wang$^{2,*}$, Siwei Cheng$^{1,*}$, Haozhe Hu$^{2,3}$}
\address{
$^{1}$University of Science and Technology of China\\
$^{2}$Institute of Digital Twin, Eastern Institute of Technology, Ningbo\\
$^{3}$Shanghai Jiao Tong University
}
\begin{document}
\ninept
\maketitle
\begingroup
\renewcommand{\thefootnote}{*}
\footnotetext{Equal contribution.}
\endgroup
\begin{abstract}
The domain of genomic language models (gLMs) has advanced rapidly, with models pretrained on diverse multi-species genomic corpora demonstrating their remarkable capabilities. While the effect of simple nucleotide-level tokenization on long-context modeling is well established, overlapping k-mer tokenization remains underexplored in causal pretraining, despite its rich neighborhood information. Here, we thoroughly revisit overlapping tokenization and introduce HGDNA, a hybrid linear-attention gLM with causal language modeling (CLM) pretraining and a shared token-prediction interface for downstream adaptation, using a species-classification auxiliary task and shared class tokens. HGDNA delivers competitive performance across sequence classification and frozen-embedding evaluations, supporting both short-range and long-range tasks with 30M parameters and 19B pretraining tokens. Code is available at \url{https://github.com/Sophie10001b/HGDNA}.
\end{abstract}


\section{Introduction}
\label{sec:Introduction}

Building on self-supervised pretraining for language~\cite{devlin2019BERT,brown2020Language} and biological sequences~\cite{lin2023Evolutionaryscale,chen2022Interpretable}, genomic language models (gLMs) explore tokenization and architectural choices to combine sensitivity to local nucleotide patterns with efficient long-context modeling under limited model and training budgets. DNABERT~\cite{ji2021DNABERT} adapts BERT to overlapping k-mers ($k\in\{3,4,5,6\}$) for regulatory sequence prediction. Nucleotide Transformer (NT)~\cite{dalla-torre2025NT} explores large-scale multi-species pretraining with non-overlapping 6-mers, while DNABERT-2~\cite{zhou2023DNABERT2} introduces byte-pair encoding (BPE)~\cite{sennrich2016Neural} to balance efficiency and downstream performance; GENA-LM~\cite{fishman2025GENALM} also uses BPE-based genomic modeling. Evo~\cite{nguyen2024Sequence}, Evo2~\cite{brixi2025Genome}, and GENERator~\cite{wu2025GENERator} further establish large-scale autoregressive genomic pretraining. For efficient long-context processing at single-nucleotide resolution, HyenaDNA~\cite{nguyen2023HyenaDNA} and Caduceus~\cite{schiff2024Caduceus} use attention-free architectures. Nevertheless, compact models still trail larger BERT-like gLMs on several downstream tasks, motivating designs that better combine local representation quality and long-range efficiency.


\begin{figure}[t]
    \centering
    \includegraphics[width=\linewidth]{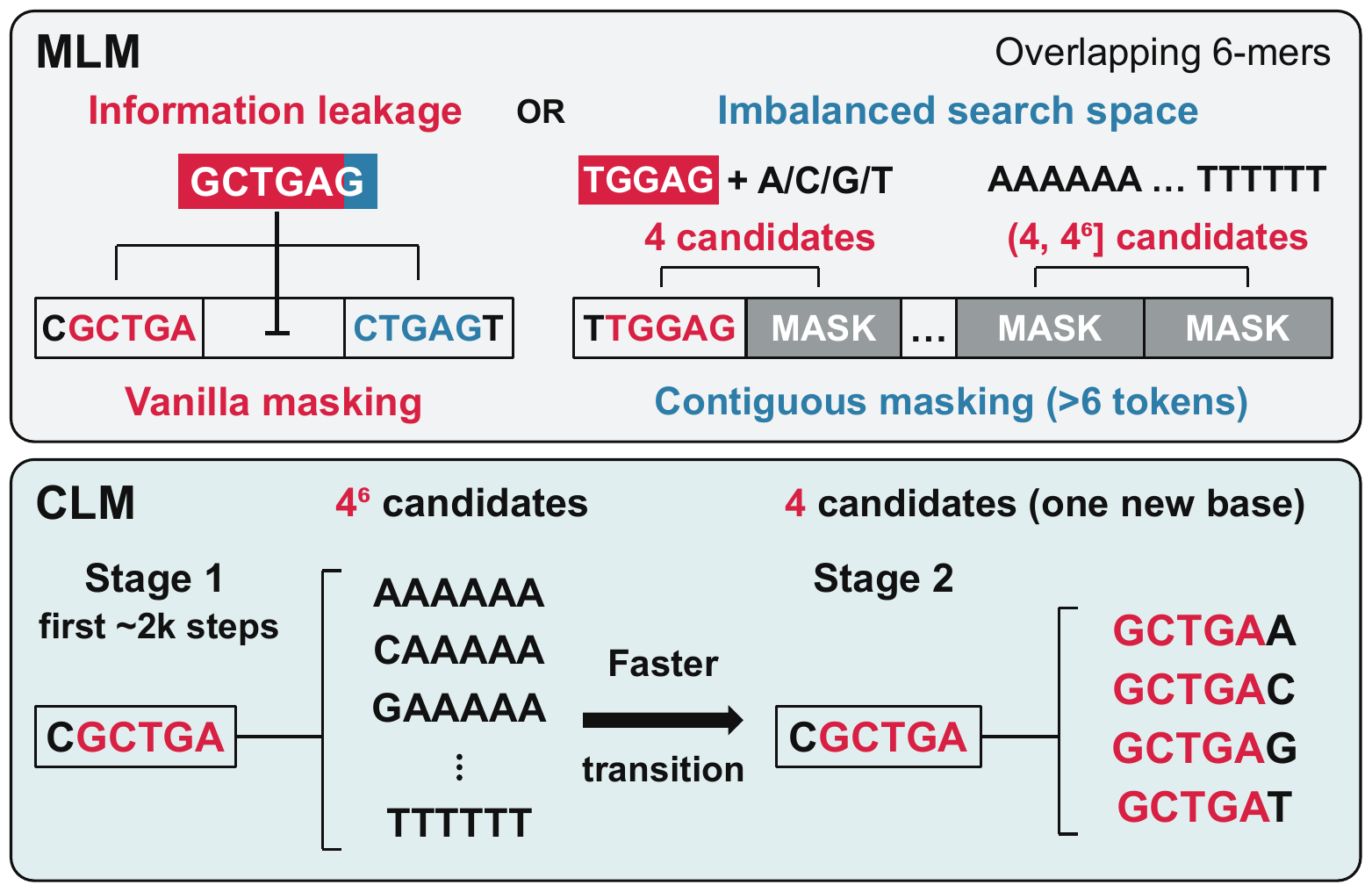}
    \caption{An overview of overlapping 6-mer pretraining in MLM and CLM objectives.}
    \label{fig:MLM_vs_CLM}
    \vspace{-10pt}
\end{figure}

Overlapping k-mer tokenization offers a promising but underexplored way to improve this balance. Unlike non-overlapping k-mers, it advances by one nucleotide per token, preserving single-nucleotide resolution while aggregating local context into a $ 4^k$-sized vocabulary. However, the shared nucleotides between adjacent tokens complicate the choice of pretraining objective. As shown in Figure~\ref{fig:MLM_vs_CLM}, masked language modeling (MLM) lets visible neighbors reveal masked content, whereas contiguous masking yields uneven recovery difficulty across masked positions. Causal language modeling (CLM) avoids access to future tokens and instead learns to exploit the overlap structure. Once it learns this structure, the preceding token determines the first $k-1$ nucleotides of the next token, reducing the effective candidate space from $4^k$ tokens to four alternatives that differ only in the new nucleotide. This transition makes CLM a natural objective for overlapping k-mer pretraining. Because overlapping tokenization does not substantially shorten a sequence, an efficient backbone remains essential. Recent advances in linear attention, particularly Gated DeltaNet~\cite{yang2024Gated}, provide data-dependent memory updates that motivate combining recurrent long-range modeling with local attention.

Towards the frontier of parameter-efficient long-context gLMs for general DNA modeling, we introduce HGDNA, a causal gLM with three main contributions.
\textbf{(1) Overlapping tokenization with CLM.}
We revisit overlapping k-mer tokenization under causal pretraining, preserving local nucleotide context while avoiding bidirectional information leakage and enabling a transition from overlap learning to single-nucleotide prediction.
\textbf{(2) A hybrid backbone for local and long-range modeling.}
We combine Gated DeltaNet with sliding window attention to capture local interactions and maintain long-range memory with linear scaling in sequence length.
\textbf{(3) Unified pretraining and downstream adaptation.}
We pretrain on gene-annotated multi-species corpus with an auxiliary species-classification objective and reuse class tokens during fine-tuning, connecting causal pretraining with sequence-level prediction.

\begin{figure*}[t]
    \centering
    \includegraphics[width=0.95\linewidth]{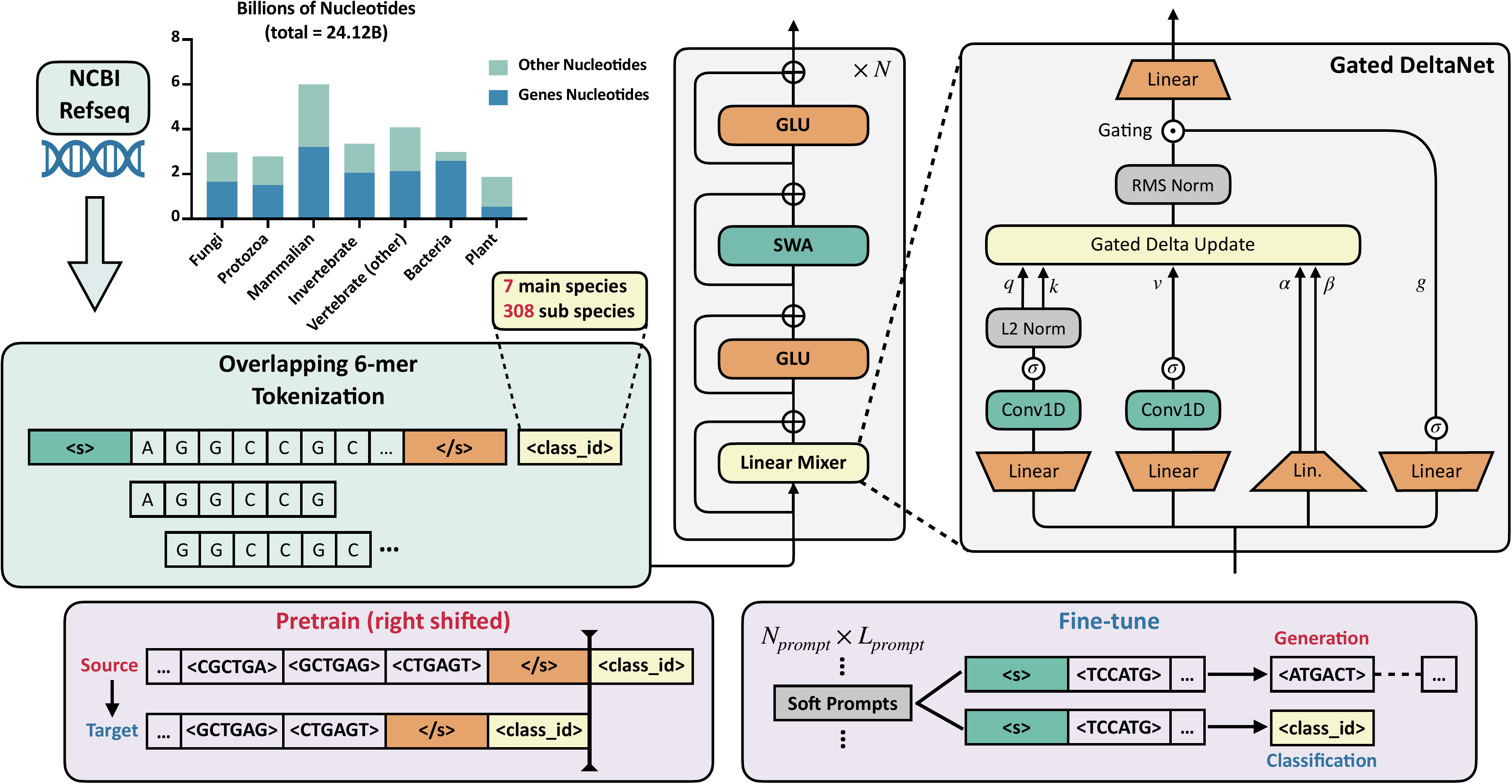}
    \caption{\textbf{Overview of HGDNA}. \textbf{Left}: gene-annotated sequences from multi-species reference genomes are tokenized into overlapping 6-mers. \textbf{Center}: the hybrid backbone combines Gated DeltaNet with sliding window attention (SWA). \textbf{Right}: the internal structure of the Gated DeltaNet module. \textbf{Bottom}: causal pretraining jointly learns sequence continuation and species-class token prediction, while downstream adaptation uses soft prompts for conditional generation and class-token prediction for classification.}
    \label{fig:overview}
    \vspace{-10pt}
\end{figure*}
\section{Method}
\label{sec:method}
As illustrated in Figure~\ref{fig:overview}, HGDNA combines overlapping tokenization with a hybrid backbone to capture local nucleotide patterns and long-range dependencies, and uses a shared token-prediction interface for pretraining and downstream adaptation.

\subsection{Overlapping Tokenization and Causal Prediction}
The vocabulary $\mathbb{V}=\mathbb{V}^{n}\cup\mathbb{V}^{c}$ contains 5,130 tokens: $\mathbb{V}^{n}$ comprises all 4,096 possible 6-mers, single-nucleotide tokens, and special tokens, while $\mathbb{V}^{c}$ comprises 1,024 reserved class tokens for pretraining and downstream classification. 

CLM predicts $u_{t+1}$ from $u_{\leq t}$. Within a DNA segment composed of A, C, G, and T, consecutive 6-mers share five nucleotides, giving the overlap-consistent candidate set
\begin{equation*}
\mathcal{C}(u_t)=
\{\operatorname{suffix}_5(u_t)\,b:
b\in\{\mathrm{A,C,G,T}\}\}.
\end{equation*}
Thus, only the newly appended nucleotide is unknown. Causal masking prevents access to future tokens containing this nucleotide (Figure~\ref{fig:MLM_vs_CLM}). This four-candidate structure is a property of the tokenized sequence; the prediction head still scores the full vocabulary during training. Stride-one tokenization largely preserves sequence length, motivating an efficient backbone.

\subsection{Hybrid Backbone with Data-Dependent Memory}
\label{sec:hybrid_backbone}
To process the token sequence efficiently, HGDNA stacks three hybrid blocks with hidden width $D=512$. Each block contains a Gated DeltaNet mixer~\cite{yang2024Gated}, a gated linear unit (GLU), sliding window attention (SWA), and another GLU, each with a residual connection. Gated DeltaNet maintains recurrent memory across the sequence, while SWA provides explicit interactions between neighboring tokens.

Within each Gated DeltaNet head, input projections and causal Conv1D produce $q_t,k_t,v_t\in\mathbb{R}^{1\times d}$; queries and keys are $\ell_2$-normalized. Input-dependent gates $\alpha_t,\beta_t\in(0,1)$ control memory decay and writing strength. The core update is
\begin{equation*}
\begin{aligned}
H_t &= \alpha_t\bigl(I-\beta_t k_t^{\top}k_t\bigr)H_{t-1}
       +\beta_t k_t^{\top}v_t,\\
\widetilde{o}_t &= q_tH_t,\qquad H_0=0,
\end{aligned}
\label{eq:gated_delta}
\end{equation*}
where $H_t\in\mathbb{R}^{d\times d}$ is the per-head memory. The delta term corrects the value retrieved by the current key after decay. As shown on the right of Figure~\ref{fig:overview}, the memory readout is RMS-normalized, modulated by an input-dependent output gate $g_t$, and projected back to the residual stream. Chunk-wise parallelization supports efficient training~\cite{yang2024Parallelizing,yang2024Gated}.

For pretraining and generation, SWA uses a causal window of 1,024 tokens and rotary position embeddings~\cite{su2024RoFormer}, with efficient attention computation~\cite{dao2023FlashAttention2}. For fixed window size and hidden width, both modules scale linearly with sequence length.

\subsection{Species-Aware Pretraining and Downstream Adaptation}

\begin{table*}[t]
  \centering
  \setlength{\tabcolsep}{2.5pt}
  \caption{\textbf{Model statistics and downstream performance on GUE, Gener, and DNA Foundation Bench.} Corpus and Tokens denote pretraining corpus size in base pairs and the number of pretraining tokens, respectively. Scores are MCC (\%), except F1 (\%) for GUE CVC; GUE Avg. averages scores over all 28 datasets. DNA Foundation Bench uses XGBoost classifiers on frozen, mean-pooled embeddings. Parentheses indicate standard deviations where reported. Best and second-best scores are shown in \textbf{bold} and \underline{underline}, respectively. $^*$ denotes input chunking for long sequences; ``--'' indicates unavailable results.}
  \label{tbl:baselines}
  \label{tbl:benchmark_GUE_NT}
  \label{tbl:benchmark_long_range}
  \aboverulesep=0ex
  \belowrulesep=0ex
  \renewcommand{\arraystretch}{1.3}

  \resizebox{\textwidth}{!}{
    \begin{tabular}{l ccc|cccccccc|cc|ccc}
      \toprule
      \multirow{2}{*}{\textbf{Models}}
      & \multicolumn{3}{c|}{\textbf{Model Statistics}}
      & \multicolumn{8}{c|}{\textbf{GUE Benchmark}}
      & \multicolumn{2}{c|}{\textbf{Gener}}
      & \multicolumn{3}{c}{\textbf{DNA Foundation Benchmark}}\\
      \cmidrule(lr){2-4}
      \cmidrule(lr){5-12}
      \cmidrule(lr){13-14}
      \cmidrule(lr){15-17}
      & \textbf{Params.} & \textbf{Corpus} & \textbf{Tokens}
      & EMP & TFM & CVC & TFH & PD & CPD & SSP & \textbf{Avg.}
      & gene & taxonomic
      & PDH & PDM & EMP\\
      \midrule

      NTv2-100M
      & 98M & 135B & 300B
      & 54.03 & 65.63 & 61.93 & 64.16 & 85.26 & 71.59 & 86.00 & 64.56
      & 62.59 (0.45) & \underline{98.02 (0.11)}$^*$
      & 83.61 (0.00) & 55.45 (0.00) & 10.30 (0.00)\\

      NTv2-500M
      & 498M & 135B & 1,000B
      & \underline{57.63} & \underline{71.59} & 63.47
      & \underline{67.36} & \textbf{91.39} & \textbf{74.74}
      & \textbf{90.49} & \underline{68.69}
      & \textbf{64.18 (0.22)} & --
      & 84.67 (0.00) & 56.77 (0.00) & \underline{11.62 (0.00)}\\

      DNABERT-2
      & 117M & 32.5B & 262B
      & 56.02 & 71.00 & \underline{71.02} & \textbf{68.95}
      & 79.56 & 69.61 & 86.15 & 66.59
      & 60.42 (0.39) & 97.44 (0.05)$^*$
      & 83.95 (0.00) & 51.44 (0.00) & 8.73 (0.00)\\
      \midrule

      HyenaDNA
      & 6M & 3.2B & 5B
      & 48.32 & 65.47 & 46.48 & 60.09 & 77.31 & 69.60 & 83.20 & 60.05
      & 56.51 (0.39) & 95.29 (0.17)
      & 84.05 (0.00) & 57.78 (0.00) & 3.58 (0.00)\\

      Caduceus-Ps
      & 8M & 3.2B & 52B
      & 53.99 & 69.74 & 68.64 & 67.22 & 75.65 & 67.05 & 83.51 & 64.46
      & 56.39 (0.43) & 96.61 (0.16)
      & \textbf{86.28 (0.00)} & \underline{60.77 (0.00)} & 3.58 (0.00)\\

      \rowcolor{lightgray}
      HGDNA
      & 30M & 13.6B & 19B
      & \textbf{68.40} & \textbf{72.33} & \textbf{72.50} & 64.78
      & \underline{86.50} & \underline{72.66} & \underline{90.02}
      & \textbf{71.77}
      & \underline{63.17 (0.51)} & \textbf{98.30 (0.30)}
      & \underline{85.47 (0.00)} & \textbf{64.26 (0.00)}
      & \textbf{13.88 (0.00)}\\
      \bottomrule
    \end{tabular}
  }
\end{table*}

\paragraph{Pretraining sequence and objective.}
Following gene-focused genomic pretraining~\cite{wu2025GENERator,brixi2025Genome}, we extract 13.69 billion bp of gene-annotated fragments from 24.12 billion bp of NCBI RefSeq genomes spanning seven taxonomic groups. We group the 887 species by shared species-name prefixes into 308 pretraining classes. We concatenate gene fragments from the same species with a delimiter $\mid$, then append the corresponding class token after the end token. Each training sequence takes the form
\begin{equation*}
\begin{aligned}
z_{1:T}=\bigl[&\langle s\rangle,\tau(g_1),\mid,\ldots,\tau(g_m),\\
              &\langle/s\rangle,c_y\bigr],
\end{aligned}
\label{eq:pretrain_sequence}
\end{equation*}
where $g_j$ is a gene fragment, $\tau$ denotes tokenization, and $c_y\in\mathbb{V}^{c}$ is the token assigned to its grouped species label $y$. Inputs $z_{1:T-1}$ predict targets $z_{2:T}$ using a single token-wise cross-entropy objective:
\begin{equation*}
\mathcal{L}_{\mathrm{pre}}
=-\frac{1}{T-1}\sum_{t=1}^{T-1}\log p_{\theta}(z_{t+1}\mid z_{1:t}).
\label{eq:pretrain_loss}
\end{equation*}
The final term predicts $c_y$ from the sequence ending at $\langle/s\rangle$, providing auxiliary sequence-level supervision through the same output head. The class token serves only as a prediction target.

\paragraph{Downstream adaptation.}
For a classification task with $C\leq1{,}024$ labels, we assign each label to a reserved class token and fine-tune the model to predict that token after $\langle/s\rangle$. Prediction uses the highest-scoring token among the $C$ task classes in a single output step~\cite{chenTokenPredictionImplicit2023}. We reuse the pretrained token-prediction head and reassign class-token meanings for each task, without adding a classification head. The input excludes the target class token.

We prepend a learnable soft prompt $P_i\in\mathbb{R}^{64\times D}$ to the sequence embeddings~\cite{lester2021softprompt,nguyen2023HyenaDNA}, and fully fine-tune the model along with the prompt parameters. For classification, all examples within a task use the same prompt $P_1$, and SWA attends to 512 neighboring tokens on each side. For conditional generation, the requested condition selects $P_i$ from a bank of prompts, and the model predicts sequence tokens autoregressively with causal SWA. Pretraining also uses causal SWA. The token-prediction interface is shared across stages, with task-specific attention masks.
\section{Experiments}

\subsection{Experimental Setup}

\paragraph{Baselines.} We compare HGDNA (30M parameters) with two groups of gLMs: (1) parameter-efficient models with comparable scales and training budgets, including HyenaDNA~\cite{nguyen2023HyenaDNA} and Caduceus-Ps~\cite{schiff2024Caduceus}; and (2) full-attention BERT-like models with larger budgets, including DNABERT-2~\cite{zhou2023DNABERT2}, NTv2~\cite{dalla-torre2025NT}, and GENA-LM~\cite{fishman2025GENALM}. HGDNA targets comparable or better performance than the latter, which use $10$--$50\times$ as many training tokens.

\paragraph{Benchmarks.} \textbf{GUE}~\cite{zhou2023DNABERT2} is a standard short-range sequence classification benchmark containing 28 datasets across 7 main tasks and 4 species, with sequence lengths from 70 bp to 1,000 bp. To explore HGDNA's potential for long-sequence understanding, we introduce the Gener Task in \textbf{GENERator}~\cite{wu2025GENERator} to assess models' performance on near 100k-bps context. Compared to task-by-task fine-tuning, zero-shot embedding presents a more efficient solution for downstream tasks with gLMs. We introduce \textbf{DNA Foundation Bench}~\cite{feng2024Benchmarking}, which contains 14 datasets covering diverse genomic prediction tasks with context lengths exceeding 2k-bps.

\begin{figure*}[htbp]
   \centering
   \includegraphics[width=0.9\linewidth]{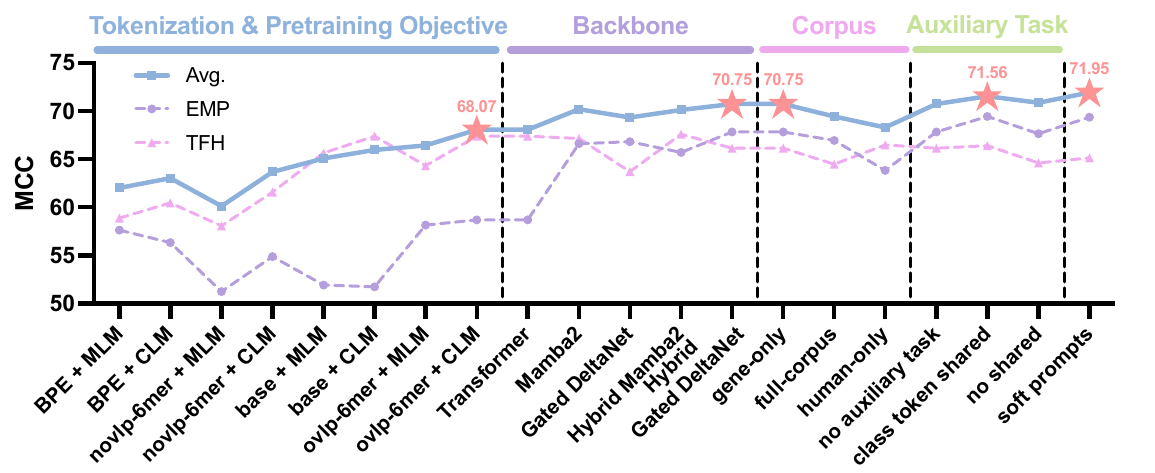}
   \caption{Ablation studies on tokenization, pretraining objectives, backbone selection, training corpus, auxiliary-task transfer, and other micro-design choices on the GUE benchmark. All models have 30M parameters and use 60\% of the standard HGDNA pretraining budget.}
   \label{fig:ablation_overview}
   \vspace{-10pt}
\end{figure*}

\paragraph{Training Details}. We perform pretraining on a single NVIDIA A100 GPU (40 GB) for 200,000 steps using AdamW with $\beta_1=0.9$, $\beta_2=0.98$, weight decay of 0.1, dropout of 0.1, and gradient clipping at 1.0. The learning rate warms up to $5\times10^{-4}$ over the first 10,000 steps and then follows cosine decay to $10^{-6}$. We linearly increase the sequence length from 64 to 2,048 over the first 40,000 steps, dynamically adjusting the batch size to accommodate at most 128 sequences and 98,304 tokens per batch. We apply reverse-complement augmentation to 50\% of sequences to increase diversity, while 10\% of sequences in each batch use randomly sampled lengths. We set the rotary embedding base to $10^6$. For the ablation experiments in Figure~\ref{fig:ablation_overview}, we use 120,000 training steps and a reduced limit of 65,536 tokens per batch.

\subsection{Main Results}

Table~\ref{tbl:baselines} highlights a favorable balance between model scale and downstream performance. With 30M parameters and 19B pretraining tokens, HGDNA achieves the highest GUE average of 71.77, exceeding NTv2-500M by 3.08 points while using only 1.9\%--7.3\% of the BERT-like baselines' reported pretraining tokens. Its largest gain is on EMP (+10.77 points over the strongest baseline), suggesting particular benefits for epigenetic mark prediction. The gains are task-dependent, however, as DNABERT-2 retains the lead on TFH.

HGDNA also maintains competitive performance when the input extends to 96 kb. On Gener's taxonomic classification task, it processes the sequence without input chunking and reaches 98.30 MCC, compared with 98.02 for NTv2-100M and 97.44 for DNABERT-2 using chunked inputs. On gene classification, HGDNA remains competitive at 63.17 MCC, 1.01 points below NTv2-500M. These results support both short-range prediction and long-context classification with a compact model.

The frozen-embedding results further indicate that the pretrained representations transfer to downstream prediction without updating the backbone. On the DNA Foundation Bench, XGBoost classifiers using HGDNA's mean-pooled embeddings outperform the strongest baselines by 3.49 points on multi-species promoter detection and 2.26 points on epigenetic mark prediction. The clearest gains are on the multi-species tasks, while human promoter detection remains slightly below Caduceus-Ps (85.47 vs.\ 86.28). This pattern complements the fine-tuning results and suggests that HGDNA learns useful features for transfer across the evaluated genomic prediction tasks.

\begin{figure}[htbp]
   \centering
   \vspace{-5pt}
   \includegraphics[width=0.9\linewidth]{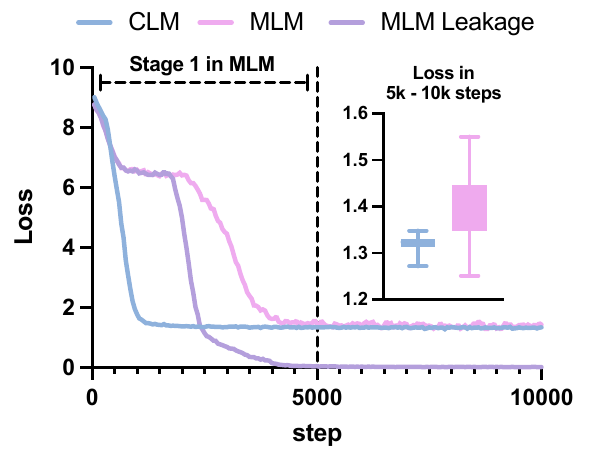}
   \caption{Pretraining loss under CLM and MLM with overlapping 6-mers. ``MLM Leakage'' uses non-contiguous masking; MLM uses contiguous masking.}
   \label{fig:pretraining_loss}
   \vspace{-10pt}
\end{figure}

\subsection{Ablation Study}
\paragraph{Macro Designs}
Figure~\ref{fig:ablation_overview} summarizes four design choices for HGDNA: tokenization and pretraining objective, backbone, training corpus, and auxiliary supervision. Relative to the DNABERT-2-like baseline (BPE + MLM), overlapping tokenization with CLM improves the average GUE score by 6.04 points, with gains on both multi-species EMP and human TFH. It also exceeds single-nucleotide tokenization by 2.97 points in average MCC. Figure~\ref{fig:pretraining_loss} shows faster initial loss reduction and lower late-stage variability under CLM than contiguous-masking MLM; the near-zero loss of ``MLM Leakage'' is consistent with target information exposed by overlapping neighbors. For the backbone, hybrid Gated DeltaNet notably benefits EMP, improving MCC by 9.12 points compared with full attention. Restricting the pretraining corpus to gene-annotated regions further improves average MCC by 1.29 points. The species-classification auxiliary task and shared class tokens jointly improve overall MCC by 0.81 points; reusing these tokens also outperforms a newly initialized [CLS] projection.
\begin{table}[!t]
   \centering
   \setlength{\tabcolsep}{2.5pt}
   \renewcommand{\arraystretch}{1.1}

   \caption{
   \textbf{Upper:} Ablation on $k$ with HGDNA without the auxiliary task.
   \textbf{Lower:} Three-layer models using only linear attention, trained from scratch for 96-kb taxonomic classification.
   }
   \label{tbl:ablation_k_and_memory}

   \begin{tabular*}{\linewidth}{@{\extracolsep{\fill}}lcccccccc@{}}
      \toprule
      $k$ & EMP & TFM & CVC & TFH & PD & CPD & SSP & \textbf{Avg.}\\
      \midrule
      3 & 58.07 & \textbf{75.23} & 69.62 & 66.18
        & 80.67 & \textbf{73.13} & 89.98 & 68.17\\
      4 & 58.92 & 72.16 & 71.89 & \textbf{67.22}
        & 82.33 & 72.48 & 88.01 & 68.23\\
      5 & 65.88 & 74.27 & 71.80 & 66.97
        & 77.02 & 70.89 & \textbf{90.74} & 70.40\\
      6 & 67.83 & 70.20 & 72.71 & 66.14
        & \textbf{83.20} & 69.89 & 88.95 & \textbf{70.75}\\
      7 & \textbf{69.98} & 70.95 & \textbf{73.28} & 66.56
        & 76.47 & 67.12 & 85.34 & 70.66\\
      \bottomrule
   \end{tabular*}

   \par\vspace{4pt}

   \begin{tabular*}{\linewidth}{@{\extracolsep{\fill}}lccc@{}}
      \toprule
      \textbf{Method}
      & \textbf{Memory Decay}
      & \textbf{KV Retrieval}
      & \textbf{MCC}\\
      \midrule
      RetNet\textsuperscript{1}
      & data-independent & -- & 76.65 (0.47)\\
      Mamba2\textsuperscript{1}
      & data-dependent & -- & 77.35 (0.41)\\
      DeltaNet\textsuperscript{1}
      & -- & $\checkmark$ & 78.56 (0.52)\\
      Gated DeltaNet\textsuperscript{1}
      & data-dependent & $\checkmark$ & \textbf{78.76 (0.45)}\\
      \bottomrule
   \end{tabular*}

   \par\vspace{3pt}

   \begin{minipage}{\linewidth}
      \raggedright
      \textsuperscript{1}using the flash-linear-attention package
      ~\cite{sun2023Retentive,dao2024Transformers,
      yang2024Parallelizing,yang2024Gated}.
   \end{minipage}
   \vspace{-10pt}
\end{table}

\paragraph{Micro Designs Tuning}
We further examine k-mer size and memory management. As shown in Table~\ref{tbl:ablation_k_and_memory}, different $k$ values can significantly affect HGDNA's performance across tasks: $k < 5$ benefits short-range TFM and CPD tasks, while larger $k$ can boost performance on longer sequences (e.g., EMP and CVC). For 96-kb taxonomic classification, memory decay and KV retrieval can improve performance, with larger gains from the Delta update.
\section{Conclusion}
We present HGDNA, combining overlapping 6-mer tokenization, a hybrid Gated DeltaNet backbone, and species-aware pretraining. With 30M parameters and 19B pretraining tokens, it achieves competitive short- and long-range classification and transferable frozen representations, supporting efficient genomic modeling.

\FloatBarrier
\clearpage
\nobalance
\section{Compliance with Ethical Standards}
This study used publicly available genomic reference sequences and benchmark datasets. No new human or animal samples were collected. No additional ethical approval was required for this study.

\bibliographystyle{IEEEbib}
\bibliography{main}
\clearpage
\nobalance
\end{document}